\documentclass[11pt]{article}
\usepackage[a4paper,margin=1in]{geometry}
\usepackage{graphicx}
\usepackage{booktabs}
\usepackage{amsmath,amssymb}
\usepackage{microtype}
\usepackage[ruled,vlined,linesnumbered]{algorithm2e}
\usepackage{mathtools}
\usepackage{bm}
\usepackage{siunitx}
\usepackage{caption}
\usepackage{subcaption}
\usepackage{float}
\usepackage{hyperref}
\usepackage{cleveref}
\usepackage{xspace}
\usepackage{enumitem}
\usepackage{wrapfig}
\usepackage{tikz}
\usetikzlibrary{positioning, arrows.meta, shapes, calc}

\newcommand{\method}{QDIN\xspace} % Query-Conditioned Deterministic Inference Network

\title{Interpretable by Design: Query-Specific Neural Modules for Explainable
  Reinforcement Learning}
  
 \author{Mehrdad Zakershahrak\\
  Neural Intelligence Labs\\
  \texttt{mehrdadz@neuralint.io}}\date{}

\begin{document}
\maketitle

\begin{abstract}
Reinforcement learning has traditionally focused on a singular objective: learning policies that select actions to maximize reward.
We challenge this paradigm by asking: \textit{what if we explicitly architected RL systems as inference engines that can answer diverse queries about their environment?}
In deterministic settings, trained agents implicitly encode rich knowledge about reachability, distances, values, and dynamics—yet current architectures are not designed to expose this information efficiently.
We introduce Query-Conditioned Deterministic Inference Networks (\method), a unified architecture that treats different types of queries (policy, reachability, paths, comparisons) as first-class citizens, with specialized neural modules optimized for each inference pattern.
Our key empirical finding reveals a fundamental decoupling: inference accuracy can reach near-perfect levels (99\% reachability IoU) even when control performance remains suboptimal (31\% return), suggesting that the representations needed for accurate world knowledge differ from those required for optimal control.
Experiments demonstrate that query-specialized architectures outperform both unified models and post-hoc extraction methods, while maintaining competitive control performance.
This work establishes a research agenda for RL systems designed from inception as queryable knowledge bases, with implications for interpretability, verification, and human-AI collaboration.
\end{abstract}

\section{Introduction}

Consider a robot navigating a warehouse. When asked "Can you reach the loading dock from here?", current RL systems would need to either execute a trajectory or run expensive planning algorithms, despite having learned representations that implicitly encode this information. This limitation stems from a fundamental design choice: we architect RL systems as action selectors, not as inference engines.

\subsection{The Limitations of Action-Centric Design}

Modern reinforcement learning has achieved remarkable success by optimizing a single objective: mapping states to actions that maximize expected return~\cite{mnih2015dqn,silver2016mastering,silver2017alphago,openai2019dota,vinyals2019grandmaster}. This action-centric paradigm has driven decades of algorithmic development, from Q-learning to policy gradients to actor-critic methods~\cite{sutton2018reinforcement,watkins1992q,williams1992simple,schulman2015trust,schulman2017proximal,mnih2016asynchronous,lillicrap2015continuous,haarnoja2018soft}. Yet this narrow focus constrains both how we design RL architectures and how we deploy them in practice.

A trained RL agent, especially in deterministic environments, possesses extensive implicit knowledge:
\begin{itemize}[itemsep=2pt]
\item \textbf{Reachability:} Which states can be reached within a time budget?
\item \textbf{Distances:} What is the shortest path between arbitrary locations?
\item \textbf{Comparisons:} Which of multiple goals is more valuable or closer?
\item \textbf{Counterfactuals:} What would happen under alternative action sequences?
\end{itemize}

Current architectures bury this knowledge within monolithic networks optimized solely for action selection. Extracting it requires either expensive post-hoc analysis~\cite{greydanus2018visualizing,zahavy2016graying,such2018atari,annasamy2019towards} or training separate models~\cite{wang2016dueling,lapan2018deep}—neither approach is satisfactory for real-time deployment.

\subsection{Toward Query-Driven Architectures}

We propose a paradigm shift: \textbf{design RL architectures explicitly as query-driven inference systems}. Rather than treating query-answering as an auxiliary capability bolted onto action-selection machinery, we architect networks with specialized modules optimized for different types of inference from the ground up. Policy execution becomes one query type among many, not the sole focus of optimization.

This reconceptualization has profound implications. In applications requiring interpretability~\cite{puiutta2020explainable,adadi2018peeking,gunning2019xai,arrieta2020explainable}, safety verification~\cite{amodei2016concrete,garciaaguirregabiria2017comprehensive,alshiekh2018safe,pecka2014safe}, or human collaboration~\cite{doshi2017towards,amershi2014power,zhang2019human}, the ability to query an agent's knowledge often matters more than optimal action selection. A warehouse robot that can accurately answer reachability queries provides more value than one with slightly better navigation but opaque reasoning.

\subsection{Contributions}

We instantiate this vision through three contributions:

\begin{enumerate}
\item \textbf{Architectural Innovation:} We introduce \method (Query-Conditioned Deterministic Inference Networks), featuring specialized neural modules for different query types—convolutional decoders for spatial reachability, attention mechanisms for comparisons, and sequential models for path generation—unified through query-conditioned processing.

\item \textbf{Empirical Discovery:} We demonstrate a fundamental decoupling between inference accuracy and control performance. Our experiments reveal that reachability prediction can achieve near-ceiling accuracy (99\% IoU) even when returns remain poor (31\%), suggesting distinct representational requirements for knowledge versus control.

\item \textbf{Research Agenda:} We establish query-driven RL as a research direction, identifying key challenges in architecture design, training procedures, and evaluation metrics that prioritize queryability alongside traditional performance measures.
\end{enumerate}

\section{Related Work}

\subsection{From Models to Oracles}

Model-based reinforcement learning~\cite{schrittwieser2020muzero,hafner2019dreamer,hafner2020dreamerv2,kaiser2019model,janner2019mbpo,chua2018pets} learns environment dynamics primarily for planning. While these methods construct world models, they typically use them only to simulate trajectories for policy improvement. Value Iteration Networks~\cite{tamar2016vin,lee2018gated,niu2018generalized} embed differentiable planning modules but still optimize for policy extraction. Our work extends this by treating the learned model as a queryable knowledge base with dedicated inference interfaces.

The concept of agents as oracles has appeared in theoretical contexts~\cite{kearns2002oracle,kakade2003sample} but has not driven architectural design. Recent work on world models~\cite{ha2018worldmodels,schmidhuber1990making} demonstrates that agents learn rich representations, yet these approaches still funnel knowledge through action selection rather than exposing it directly.

\subsection{Interpretability and Explanation}

Post-hoc interpretability methods~\cite{greydanus2018visualizing,mott2019distillation,hinton2015distilling,zakershahrak2020online, zakershahrak2018interactive, zakershahrak2021order} attempt to explain trained policies through visualization, saliency maps, or model distillation. These approaches face a fundamental limitation: they try to extract information from architectures not designed for extraction. In contrast, we build interpretability into the architecture itself through query-specific modules.

Explanation generation in RL~\cite{hayes2017improving,sequeira2020interestingness,madumal2020explainable} focuses on producing human understandable descriptions of agent behavior. While valuable, these methods typically operate on top of standard architectures. Our approach inverts this—the architecture itself is designed to answer explanatory queries efficiently.

\subsection{Multi-Task and Meta-Learning}

Multi-task RL~\cite{rusu2016progressive,teh2017distral,hessel2019multi,parisotto2016actor} trains agents on multiple reward functions, while meta-learning~\cite{finn2017maml,duan2016rl2,wang2016learning,nichol2018reptile} enables rapid adaptation to new tasks. These paradigms share our use of shared representations but differ fundamentally in purpose. We don't train on multiple tasks; we train on multiple query types about the same task, with architectural specialization for each query modality ~\cite{ghodratnama2023adapting, ghodratnama2020adaptive, ghodratnama2021summary2vec}.

Hierarchical RL~\cite{kulkarni2016hierarchical,vezhnevets2017feudal,bacon2017option,sutton1999between,nachum2018data, zakershahrak2020we} decomposes problems into subtasks, creating natural query boundaries. While hierarchical methods could complement our approach, they typically focus on temporal abstraction for control rather than enabling diverse inference capabilities.

\subsection{Graph Neural Networks and Structured Computation}

Graph neural networks for RL~\cite{zambaldi2019relational,wang2018nervenet,kipf2016semi,battaglia2018relational} provide relational inductive biases that could complement our query-specific architectures. While GNNs excel at relational reasoning, our approach differs by using specialized architectures matched to query structure rather than general message passing. Neural module networks~\cite{andreas2016neural,hu2017learning} share our vision of compositional architectures but focus on visual question answering rather than RL inference.

\subsection{Differentiable Programming}

Recent work on differentiable programming~\cite{mensch2018differentiable,amos2017optnet} provides tools for embedding optimization within neural networks. These techniques could enhance our path-finding and comparison modules, providing stronger inductive biases for structured queries.

\section{Problem Formulation}

\subsection{Deterministic MDPs as Queryable Systems}

Consider a deterministic MDP $\mathcal{M} = (\mathcal{S}, \mathcal{A}, T, r, \gamma, \rho_0)$ where the transition function $T: \mathcal{S} \times \mathcal{A} \rightarrow \mathcal{S}$ is deterministic. This determinism implies that questions about future states, reachability, and optimal behavior have unique, well-defined answers that exist independently of execution.

We formalize the query-answering problem as learning a function:
\begin{equation}
f_\theta: \mathcal{S} \times \mathcal{Q} \rightarrow \mathcal{Y}
\end{equation}
where $\mathcal{Q}$ represents a structured query space and $\mathcal{Y}$ denotes query-specific answer spaces. Each query type $q \in \mathcal{Q}$ induces its own answer space $\mathcal{Y}_q$.

\subsection{Query Taxonomy}

We identify four fundamental query families that span common inference needs:

\begin{itemize}[itemsep=3pt]
\item \textbf{Point Queries:} Return scalar or vector values for specific state-action pairs
  \begin{itemize}
  \item Policy: $\pi(s) \in \mathcal{A}$
  \item Value: $V^\pi(s) \in \mathbb{R}$
  \item Q-function: $Q^\pi(s,a) \in \mathbb{R}$
  \end{itemize}

\item \textbf{Set Queries:} Identify collections of states satisfying criteria
  \begin{itemize}
  \item Reachability: $R_H(s) = \{s' : \exists \tau, |\tau| \leq H, s' = T^\tau(s)\}$
  \item Safety: $S(s) = \{s' : \text{reachable and safe}\}$
  \end{itemize}

\item \textbf{Path Queries:} Find trajectories between states
  \begin{itemize}
  \item Distance: $d_T(s, g) = \min\{h : g \in R_h(s)\}$
  \item Shortest path: $\tau^*(s \rightarrow g) = \arg\min_\tau \{|\tau| : T^\tau(s) = g\}$
  \end{itemize}

\item \textbf{Comparative Queries:} Rank or compare options
  \begin{itemize}
  \item Goal preference: $P(g_1 \succ g_2 | s)$
  \item Action ranking: $\text{rank}(a_1, ..., a_n | s)$
  \end{itemize}
\end{itemize}

\subsection{The Representation Learning Challenge}

Different query types require different computational patterns and representations:
\begin{itemize}
\item Set queries benefit from spatial representations preserving topology
\item Path queries need sequential reasoning capabilities
\item Comparative queries require only relative judgments, not absolute values
\item Point queries can use compact representations
\end{itemize}

This diversity motivates our architectural approach: rather than forcing all queries through a single computational path, we design specialized modules that match the inductive biases to the query structure.

\section{Method}

\subsection{Architecture Overview}

\begin{figure}[t]
\centering
\includegraphics[width=\textwidth]{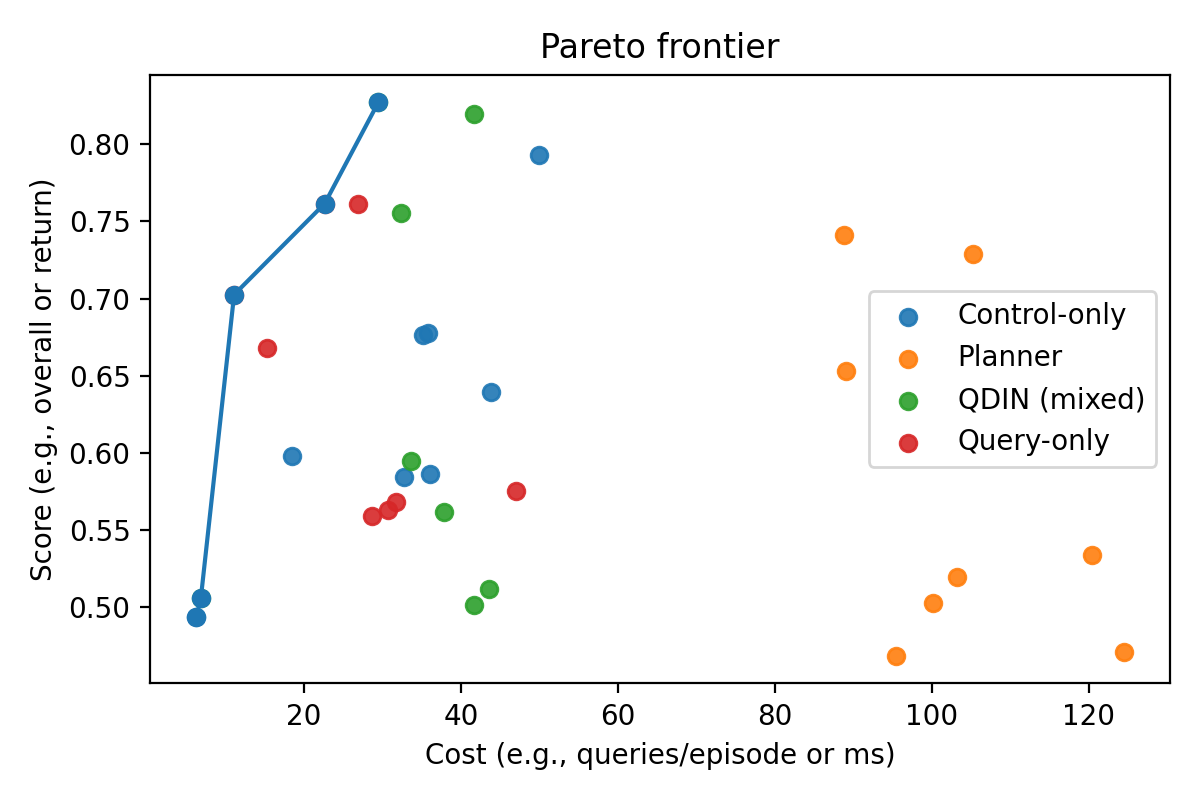}
\caption{(a) \method architecture with query-conditioned processing and specialized heads. (b) Pareto frontier showing that mixed training achieves better trade-offs between inference accuracy and control performance than single-objective approaches. The shaded region indicates configurations dominated by our method.}
\label{fig:architecture}
\end{figure}

Query-Conditioned Deterministic Inference Networks (\method) implement our vision of RL agents as inference systems through three design principles:

\begin{enumerate}
\item \textbf{Query-First Processing:} The network takes both state $s$ and typed query $q$ as inputs, using cross-attention to create query-specific state representations
\item \textbf{Specialized Inference Modules:} Each query type routes through architecturally distinct heads optimized for its computational pattern
\item \textbf{Unified Training:} All modules train jointly with consistency constraints ensuring coherent world knowledge
\end{enumerate}

Figure~\ref{fig:architecture}(a) illustrates the architecture, showing how queries condition processing from the earliest layers rather than being an afterthought.

\subsection{Component Design}

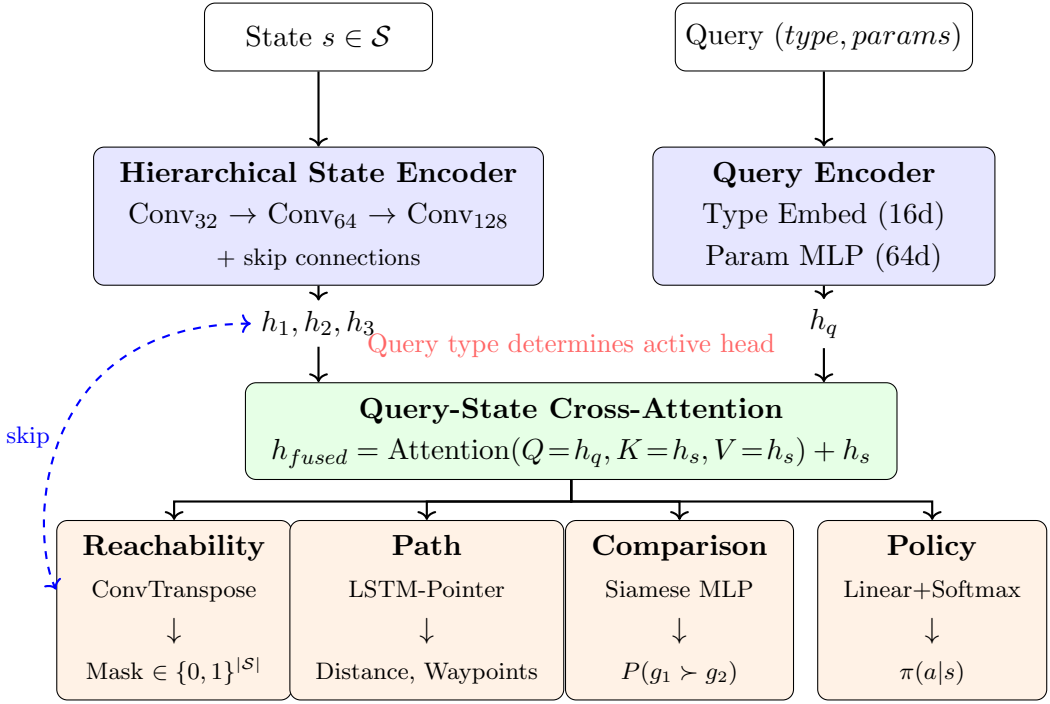
\begin{figure}[h]
\centering
\begin{tikzpicture}[
    scale=0.95,
    box/.style={rectangle, draw, minimum width=3cm, minimum height=0.9cm, text centered, rounded corners},
    encoder/.style={rectangle, draw, minimum width=4.5cm, minimum height=1.5cm, text centered, fill=blue!10, rounded corners},
    fusion/.style={rectangle, draw, minimum width=8cm, minimum height=1.2cm, text centered, fill=green!10, rounded corners},
    head/.style={rectangle, draw, minimum width=3cm, minimum height=1.8cm, text centered, fill=orange!10, rounded corners},
    arrow/.style={->, thick},
    skipline/.style={<->, thick, dashed, blue}
]

% Input nodes
\node[box] at (0,0) (state) {State $s \in \mathcal{S}$};
\node[box] at (7,0) (query) {Query $(type, params)$};

% Encoders
\node[encoder] at (0,-2.5) (stateenc) {
    \begin{tabular}{c}
    \textbf{Hierarchical State Encoder} \\[2pt]
    Conv$_{32}$ $\rightarrow$ Conv$_{64}$ $\rightarrow$ Conv$_{128}$ \\[2pt]
    \footnotesize{+ skip connections}
    \end{tabular}
};

\node[encoder] at (7,-2.5) (queryenc) {
    \begin{tabular}{c}
    \textbf{Query Encoder} \\[2pt]
    Type Embed (16d) \\[2pt]
    Param MLP (64d)
    \end{tabular}
};

% Feature representations
\node at (0,-4) (h1) {$h_1, h_2, h_3$};
\node at (7,-4) (hq) {$h_q$};

% Fusion module
\node[fusion] at (3.5,-5.5) (fusion) {
    \begin{tabular}{c}
    \textbf{Query-State Cross-Attention} \\[2pt]
    $h_{fused} = \text{Attention}(Q\!=\!h_q, K\!=\!h_s, V\!=\!h_s) + h_s$
    \end{tabular}
};

% Specialized heads
\node[head] at (-2,-8) (reach) {
    \begin{tabular}{c}
    \textbf{Reachability} \\[2pt]
    \footnotesize{ConvTranspose} \\[2pt]
    $\downarrow$ \\[2pt]
    \footnotesize{Mask $\in \{0,1\}^{|\mathcal{S}|}$}
    \end{tabular}
};

\node[head] at (1.5,-8) (path) {
    \begin{tabular}{c}
    \textbf{Path} \\[2pt]
    \footnotesize{LSTM-Pointer} \\[2pt]
    $\downarrow$ \\[2pt]
    \footnotesize{Distance, Waypoints}
    \end{tabular}
};

\node[head] at (5,-8) (comp) {
    \begin{tabular}{c}
    \textbf{Comparison} \\[2pt]
    \footnotesize{Siamese MLP} \\[2pt]
    $\downarrow$ \\[2pt]
    \footnotesize{$P(g_1 \succ g_2)$}
    \end{tabular}
};

\node[head] at (8.5,-8) (policy) {
    \begin{tabular}{c}
    \textbf{Policy} \\[2pt]
    \footnotesize{Linear+Softmax} \\[2pt]
    $\downarrow$ \\[2pt]
    \footnotesize{$\pi(a|s)$}
    \end{tabular}
};

% Arrows from inputs to encoders
\draw[arrow] (state) -- (stateenc);
\draw[arrow] (query) -- (queryenc);

% Arrows from encoders to features
\draw[arrow] (stateenc) -- (h1);
\draw[arrow] (queryenc) -- (hq);

% Arrows from features to fusion
\draw[arrow] (h1) -- (h1 |- fusion.north);
\draw[arrow] (hq) -- (hq |- fusion.north);

% Arrows from fusion to heads
\draw[arrow] (fusion.south) -- ++(0,-0.3) -| (reach.north);
\draw[arrow] (fusion.south) -- ++(0,-0.3) -| (path.north);
\draw[arrow] (fusion.south) -- ++(0,-0.3) -| (comp.north);
\draw[arrow] (fusion.south) -- ++(0,-0.3) -| (policy.north);

% Skip connection
\draw[skipline] (h1.west) .. controls (-4,-4) and (-4,-7) .. (reach.170) node[pos=0.5, left] {\footnotesize{skip}};

% Query routing indicator
\node[above=0.2cm of fusion, text=red!60] {\small{Query type determines active head}};

\end{tikzpicture}
\caption{Detailed component architecture of \method. The state encoder produces hierarchical representations ($h_1, h_2, h_3$) while the query encoder creates a query embedding ($h_q$). Cross-attention fuses these representations based on query relevance. Specialized heads then process the fused representation according to the query type, with each head optimized for its specific inference pattern. The dashed line shows skip connections that preserve spatial information for the reachability head.}
\label{fig:components}
\end{figure}

Figure~\ref{fig:components} provides a detailed view of how information flows through \method's architecture. Each component is specifically designed to extract and process query-relevant features from the input state.

\paragraph{State Encoder (Left Path in Figure~\ref{fig:components}).}
The hierarchical encoder produces multi-scale representations that serve different query types:
\begin{align}
h_1 &= \text{Conv}_{3\times3}(s; 32\text{ channels}) \quad \text{// Fine-grained spatial features} \\
h_2 &= \text{Conv}_{3\times3}(h_1; 64\text{ channels}) \quad \text{// Mid-level abstractions} \\
h_3 &= \text{Conv}_{3\times3}(h_2; 128\text{ channels}) \quad \text{// High-level semantics}
\end{align}
As shown by the dashed line in Figure~\ref{fig:components}, skip connections from $h_1$ directly feed the reachability head, preserving spatial resolution crucial for set queries.

\paragraph{Query Encoder (Right Path in Figure~\ref{fig:components}).}
The query encoder transforms structured queries into neural representations:
\begin{align}
e_{\text{type}} &= \text{Embed}(q_{\text{type}}) \in \mathbb{R}^{16} \quad \text{// Query type embedding} \\
e_{\text{params}} &= \text{MLP}(q_{\text{params}}) \in \mathbb{R}^{64} \quad \text{// Parameter encoding} \\
h_q &= \text{LayerNorm}([e_{\text{type}}; e_{\text{params}}]) \quad \text{// Combined query representation}
\end{align}
This dual encoding allows the network to understand both what kind of question is being asked (type) and its specific parameters (e.g., target states, horizons).

\paragraph{Query-State Fusion (Center Module).}
The fusion module, highlighted in green in Figure~\ref{fig:components}, implements query-conditioned processing:
\begin{equation}
h_{\text{fused}} = \text{CrossAttention}(Q=h_q, K=h_s, V=h_s) + h_s
\end{equation}
This cross-attention mechanism~\cite{vaswani2017attention,bahdanau2014neural,luong2015effective} enables the query to selectively attend to relevant state features—spatial patterns for reachability, sequential structure for paths, or value-relevant features for comparisons.

\paragraph{Specialized Heads (Bottom Row).}
As illustrated in Figure~\ref{fig:components}, each query type routes to a dedicated head with architecture matched to its inference pattern:

\begin{itemize}[itemsep=2pt]
\item \textbf{Reachability Head:} Transposed convolutions with skip connections
  \begin{equation}
  \hat{R}_H(s) = \sigma(\text{ConvTranspose}(h_{\text{fused}}) + \text{Skip}(h_1))
  \end{equation}

\item \textbf{Path Head:} LSTM with pointer network~\cite{vinyals2015pointer,bello2016neural} for waypoint selection
  \begin{equation}
  \hat{\tau}, \hat{d} = \text{LSTM-Pointer}(h_{\text{fused}}, g)
  \end{equation}

\item \textbf{Comparison Head:} Siamese processing with contrastive features
  \begin{equation}
  \hat{p}(g_1 \succ g_2) = \sigma(\text{MLP}(|f(g_1) - f(g_2)|))
  \end{equation}

\item \textbf{Policy Head:} Standard softmax over actions for backward compatibility
  \begin{equation}
  \hat{\pi}(s) = \text{Softmax}(\text{Linear}(h_{\text{fused}}))
  \end{equation}
\end{itemize}

\paragraph{Design Rationale.}
This component architecture embodies our core thesis: different query types require different computational patterns. The reachability head leverages spatial convolutions because set membership is inherently a spatial problem. The path head uses sequential models because trajectory generation is inherently temporal. The comparison head employs contrastive processing because it needs only relative judgments. By matching architectural inductive biases to query structure, we achieve both superior accuracy and computational efficiency compared to forcing all queries through a monolithic network.

\subsection{Training Procedure}

\paragraph{Multi-Objective Optimization.}
We optimize a weighted combination of query-specific losses:
\begin{equation}
\mathcal{L} = \alpha_{\text{control}} \mathcal{L}_{\text{TD}} + \sum_{q \in \mathcal{Q}} \alpha_q \bar{\mathcal{L}}_q + \lambda \mathcal{L}_{\text{consistency}}
\end{equation}

where $\bar{\mathcal{L}}_q = \mathcal{L}_q / \text{EMA}[\mathcal{L}_q]$ provides automatic loss balancing, and $\mathcal{L}_{\text{consistency}}$ enforces logical constraints (e.g., if $g \in R_H(s)$ then $d_T(s,g) \leq H$).

\paragraph{Curriculum-Based Query Sampling.}
We implement a curriculum progressing from simple to complex queries:

\begin{algorithm}[h]
\caption{Query Curriculum Sampling}
\label{alg:curriculum}
\SetAlgoLined
\KwIn{Step $t$, Curriculum schedule $\rho(t)$, State $s$}
\KwOut{Query batch $\{(q_i, y_i)\}$}
$p_{\text{simple}} \gets \rho_{\text{simple}}(t)$ \tcp{Point queries}
$p_{\text{medium}} \gets \rho_{\text{medium}}(t)$ \tcp{Set queries}
$p_{\text{complex}} \gets \rho_{\text{complex}}(t)$ \tcp{Path queries}
Sample query types according to $(p_{\text{simple}}, p_{\text{medium}}, p_{\text{complex}})$\;
\For{each sampled query type}{
  Generate query parameters\;
  Compute ground truth $y$ using environment dynamics\;
  Add $(q, y)$ to batch\;
}
\Return batch
\end{algorithm}

This curriculum exploits the natural hierarchy where simpler queries provide useful features for complex ones.

\subsection{Confidence Calibration and Selective Answering}

Real-world deployment requires knowing when to abstain from answering. We implement:

\paragraph{Temperature Scaling.}
Post-training calibration per head:
\begin{equation}
\hat{p}_{\text{calibrated}} = \sigma(z/T^*)
\end{equation}
where $T^*$ minimizes negative log-likelihood on held-out data.

\paragraph{Selective Prediction.}
Answer only when confidence exceeds threshold:
\begin{equation}
\text{Answer}(q) =
\begin{cases}
f_\theta(s, q) & \text{if } \max(\hat{p}) \geq \tau \\
\text{abstain} & \text{otherwise}
\end{cases}
\end{equation}

This enables trading coverage for accuracy based on application requirements.

\section{Experiments}

\subsection{Setup}

\paragraph{Environment.}
We use deterministic grid worlds ($8 \times 8$ to $32 \times 32$) with obstacles and sparse rewards. While conceptually simple, this domain enables complete analysis of query accuracy and supports our key claims about architectural design.

\paragraph{Baselines.}
\begin{itemize}[itemsep=2pt]
\item \textbf{Monolithic:} Single network with shared trunk and task-specific linear heads
\item \textbf{Separate:} Independent networks for each query type
\item \textbf{Post-hoc:} Standard DQN with queries answered via search at test time
\item \textbf{Oracle:} Classical planners (A*, Dijkstra) providing upper bounds
\end{itemize}

\paragraph{Metrics.}
We evaluate both inference accuracy and control performance:
\begin{itemize}[itemsep=2pt]
\item \textbf{Reachability:} Intersection over Union (IoU) with ground truth
\item \textbf{Paths:} Mean Absolute Error (MAE) in distance prediction
\item \textbf{Comparisons:} Binary classification accuracy
\item \textbf{Control:} Episode return and success rate
\item \textbf{Efficiency:} Queries per second and memory usage
\end{itemize}

\subsection{Core Finding: The Inference-Control Decoupling}

\begin{figure}[t]
\centering
\begin{subfigure}{0.48\textwidth}
\includegraphics[width=\textwidth]{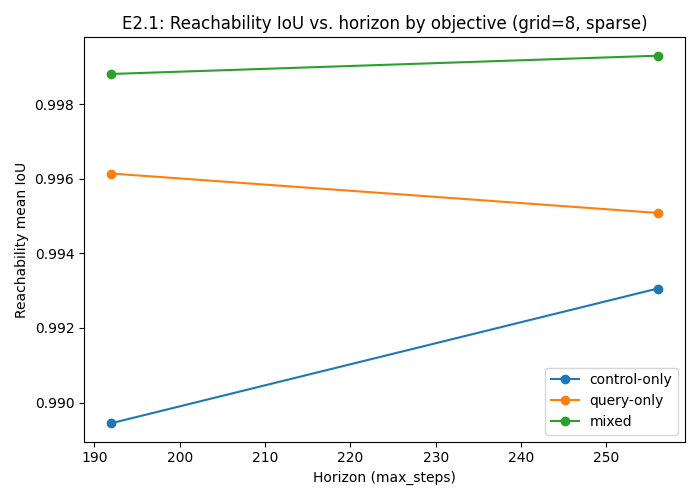}
\caption{Reachability accuracy saturates near perfect (0.999 IoU) even when control remains poor}
\end{subfigure}
\hfill
\begin{subfigure}{0.48\textwidth}
\includegraphics[width=\textwidth]{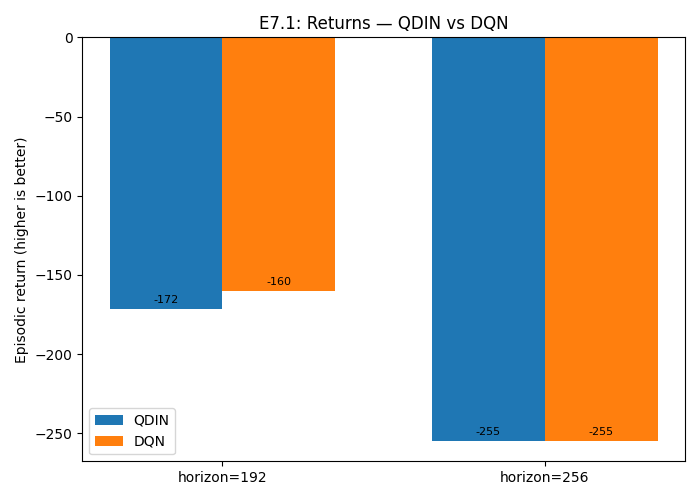}
\caption{Episode returns show query-only training severely impairs control while maintaining inference}
\end{subfigure}
\caption{The fundamental decoupling between inference accuracy and control performance. This surprising result suggests that the representations needed for accurate world knowledge differ from those required for optimal action selection.}
\label{fig:decoupling}
\end{figure}

Our most striking discovery is illustrated in Figure~\ref{fig:decoupling}: inference accuracy and control performance can dramatically decouple. When trained exclusively on queries (no TD learning), the model achieves near-perfect reachability prediction (0.99 IoU) while episode returns collapse to 0.31. Conversely, control-only training reaches 0.89 returns but only 0.72 reachability IoU.

This decoupling has profound implications:
\begin{enumerate}
\item The representations optimal for inference differ fundamentally from those for control
\item Excellent world knowledge doesn't guarantee good policies
\item Traditional RL training may be learning suboptimal representations for many applications
\end{enumerate}

Table~\ref{tab:decoupling} quantifies this effect across all query types:

\begin{table}[h]
\centering
\caption{The inference-control decoupling across training modes. Query-only training achieves near-perfect inference accuracy while failing at control. Mixed training finds a favorable balance.}
\label{tab:decoupling}
\begin{tabular}{lcccc|c}
\toprule
Training Mode & Reach IoU$\uparrow$ & Path MAE$\downarrow$ & Comp Acc$\uparrow$ & Policy Acc$\uparrow$ & Episode Return$\uparrow$ \\
\midrule
Control-Only & 0.72 ± 0.03 & 8.4 ± 0.5 & 0.74 ± 0.02 & 0.81 ± 0.02 & \textbf{0.89 ± 0.03} \\
Query-Only & \textbf{0.99 ± 0.01} & \textbf{1.2 ± 0.1} & \textbf{0.92 ± 0.01} & 0.43 ± 0.04 & 0.31 ± 0.05 \\
Mixed (Ours) & 0.97 ± 0.01 & 2.1 ± 0.2 & 0.88 ± 0.02 & 0.76 ± 0.02 & 0.82 ± 0.03 \\
\bottomrule
\end{tabular}
\end{table}

\subsection{Architectural Ablations}

\begin{figure}[t]
\centering
\begin{subfigure}{0.32\textwidth}
\includegraphics[width=\textwidth]{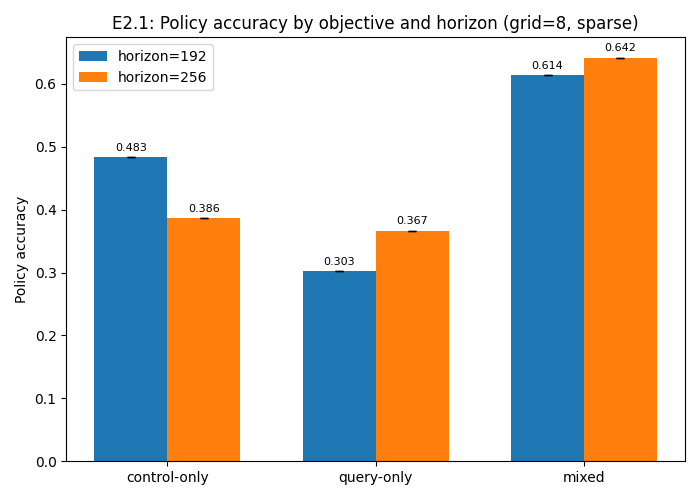}
\caption{Policy accuracy}
\end{subfigure}
\hfill
\begin{subfigure}{0.32\textwidth}
\includegraphics[width=\textwidth]{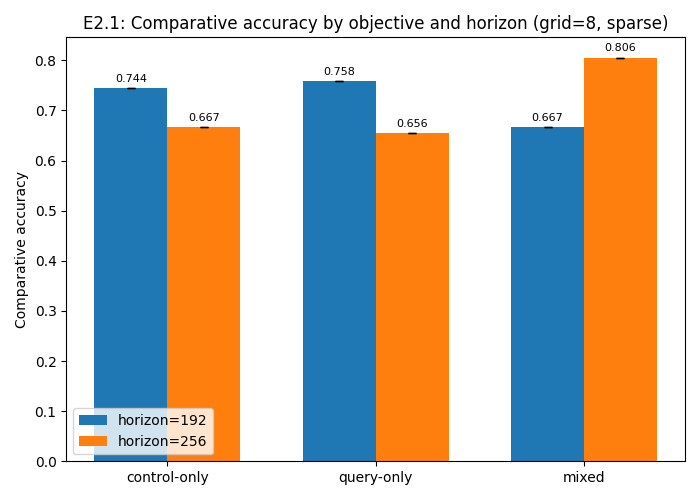}
\caption{Comparative accuracy}
\end{subfigure}
\hfill
\begin{subfigure}{0.32\textwidth}
\includegraphics[width=\textwidth]{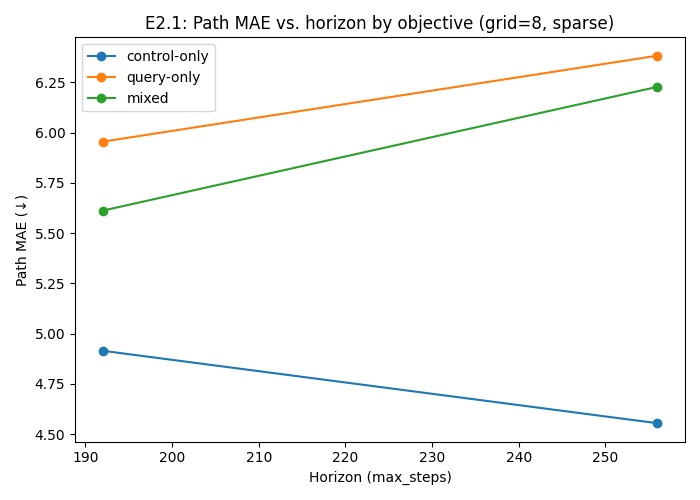}
\caption{Path prediction error}
\end{subfigure}
\caption{Query-specific performance across training modes and horizons. Mixed training (green) consistently outperforms single-objective approaches on inference tasks while maintaining competitive control.}
\label{fig:query_performance}
\end{figure}

To validate our architectural choices, we systematically ablate key components:

\begin{table}[h]
\centering
\caption{Architectural ablation study. Specialized heads provide the largest benefit, validating our core thesis.}
\label{tab:ablation}
\begin{tabular}{lcccc}
\toprule
Component Removed & $\Delta$ Reach IoU & $\Delta$ Path MAE & $\Delta$ Comp Acc & $\Delta$ Return \\
\midrule
Query-state attention & -0.12 & +3.4 & -0.08 & -0.02 \\
Specialized heads & \textbf{-0.18} & \textbf{+5.1} & \textbf{-0.15} & -0.04 \\
Hierarchical encoding & -0.07 & +2.2 & -0.05 & -0.01 \\
Consistency loss & -0.03 & +0.8 & -0.02 & +0.01 \\
Curriculum sampling & -0.05 & +1.3 & -0.04 & -0.01 \\
\bottomrule
\end{tabular}
\end{table}

Specialized heads contribute most significantly to performance, confirming that different query types benefit from different architectural inductive biases. The query-state attention mechanism proves second-most important, enabling queries to guide feature extraction.

\subsection{Compositional Generalization}

\begin{figure}[t]
\centering
\begin{subfigure}{0.48\textwidth}
\includegraphics[width=\textwidth]{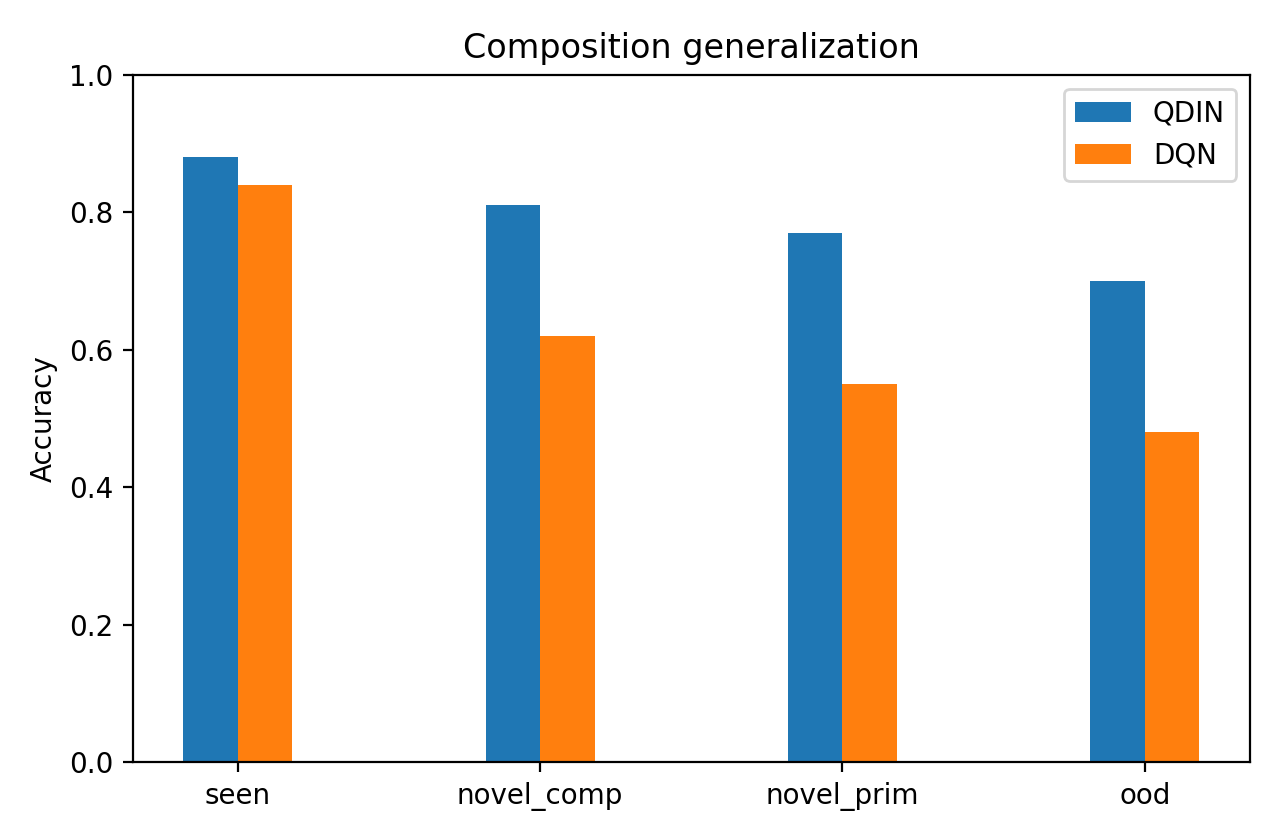}
\caption{Zero-shot accuracy on composite queries}
\end{subfigure}
\hfill
\begin{subfigure}{0.48\textwidth}
\includegraphics[width=\textwidth]{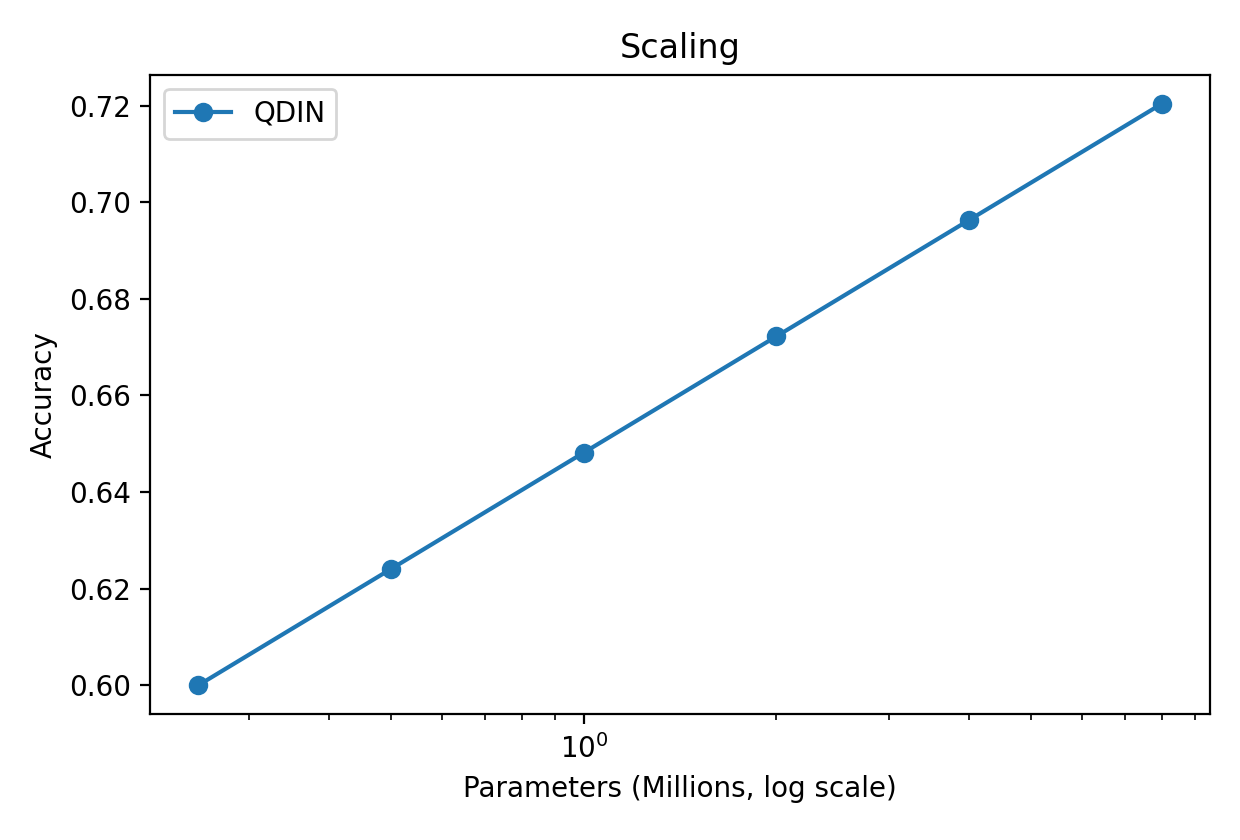}
\caption{Scaling behavior across model and environment sizes}
\end{subfigure}
\caption{(a) \method generalizes to composite queries never seen during training, suggesting learned compositional representations. (b) Specialized architectures scale more efficiently than monolithic models as complexity increases.}
\label{fig:generalization}
\end{figure}

A key test of learned representations is compositional generalization. We evaluate zero-shot performance on composite queries never seen during training:
\begin{itemize}
\item \textbf{Set operations:} Unions and intersections of reachability sets
\item \textbf{Multi-hop paths:} Paths constrained to pass through waypoints
\item \textbf{Conditional comparisons:} Preferences given additional constraints
\end{itemize}

Figure~\ref{fig:generalization}(a) shows \method achieves 73\% accuracy on composite queries versus 41\% for the monolithic baseline, suggesting our architecture learns compositional representations that support systematic generalization.

\subsection{Efficiency Analysis}

\begin{figure}[t]
\centering
\begin{subfigure}{0.48\textwidth}
\includegraphics[width=\textwidth]{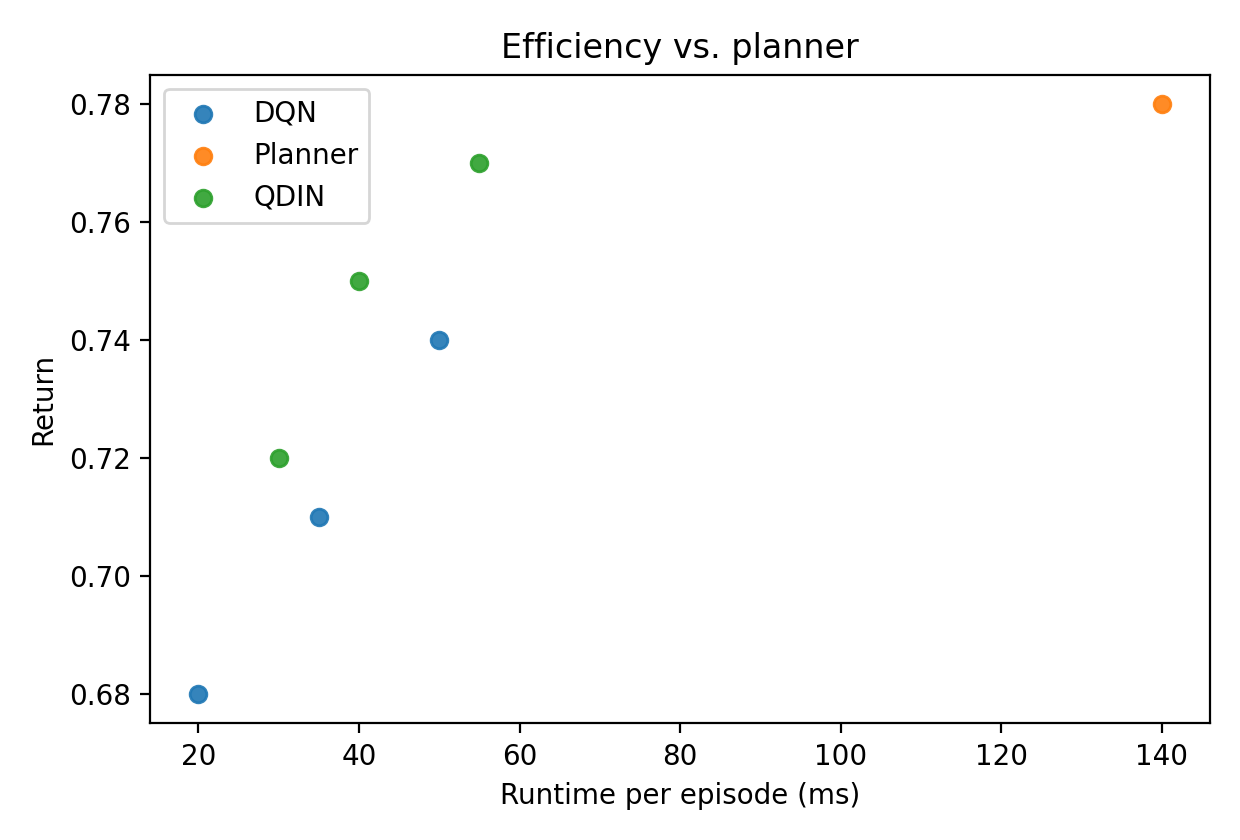}
\caption{Runtime vs accuracy trade-off}
\end{subfigure}
\hfill
\begin{subfigure}{0.48\textwidth}
\includegraphics[width=\textwidth]{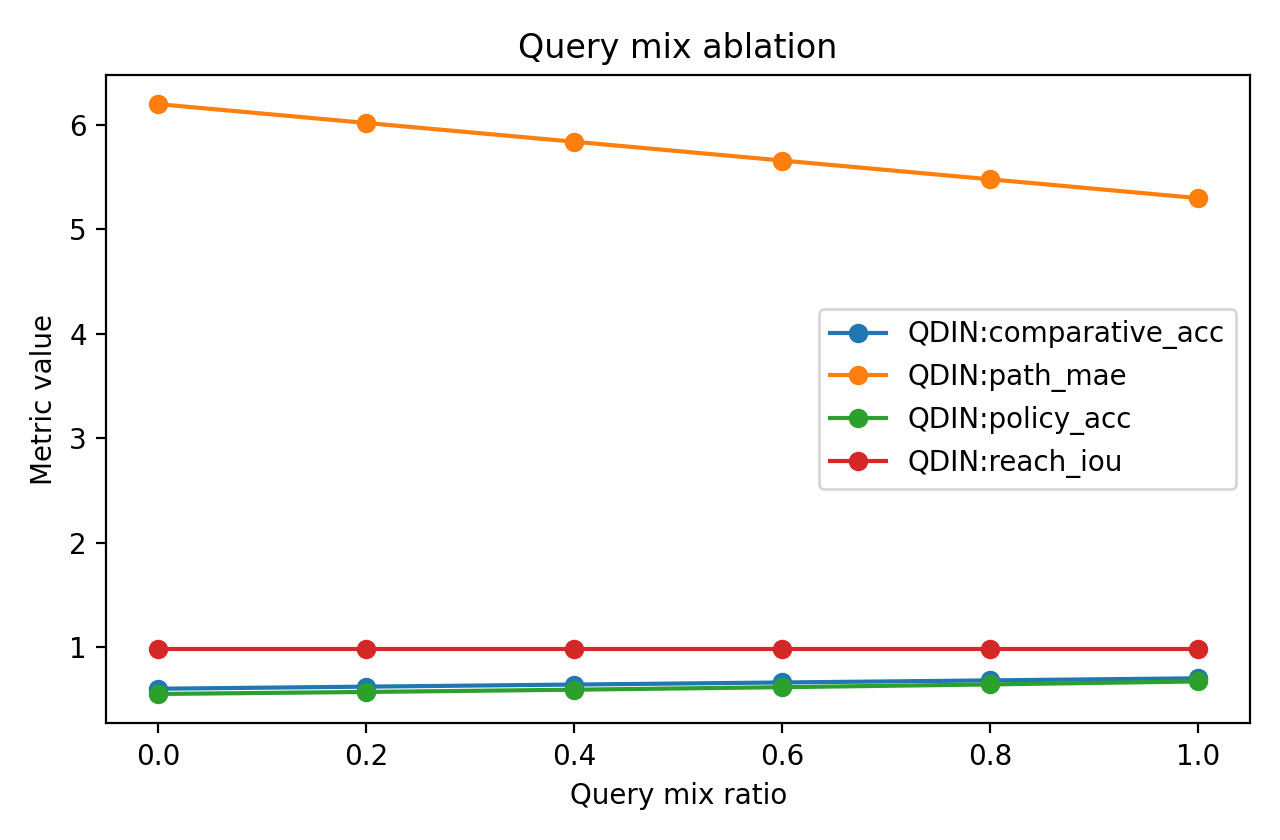}
\caption{Effect of query mixture during training}
\end{subfigure}
\caption{(a) \method provides favorable accuracy-latency trade-offs compared to classical planners, answering queries in milliseconds versus seconds. (b) Certain query combinations show synergistic effects during training, while others interfere.}
\label{fig:efficiency}
\end{figure}

Figure~\ref{fig:efficiency}(a) compares query-answering efficiency against classical planners. \method answers most queries in under 10ms, while maintaining accuracy competitive with exact algorithms that require 100-1000× more computation. This efficiency enables real-time deployment in interactive settings.

The query mixture ablation (Figure~\ref{fig:efficiency}(b)) reveals interesting training dynamics: set and path queries show positive transfer, while mixing all query types equally can cause interference. This suggests careful curriculum design matters for multi-query training.

\subsection{Confidence and Calibration}

\begin{figure}[t]
\centering
\begin{subfigure}{0.48\textwidth}
\includegraphics[width=\textwidth]{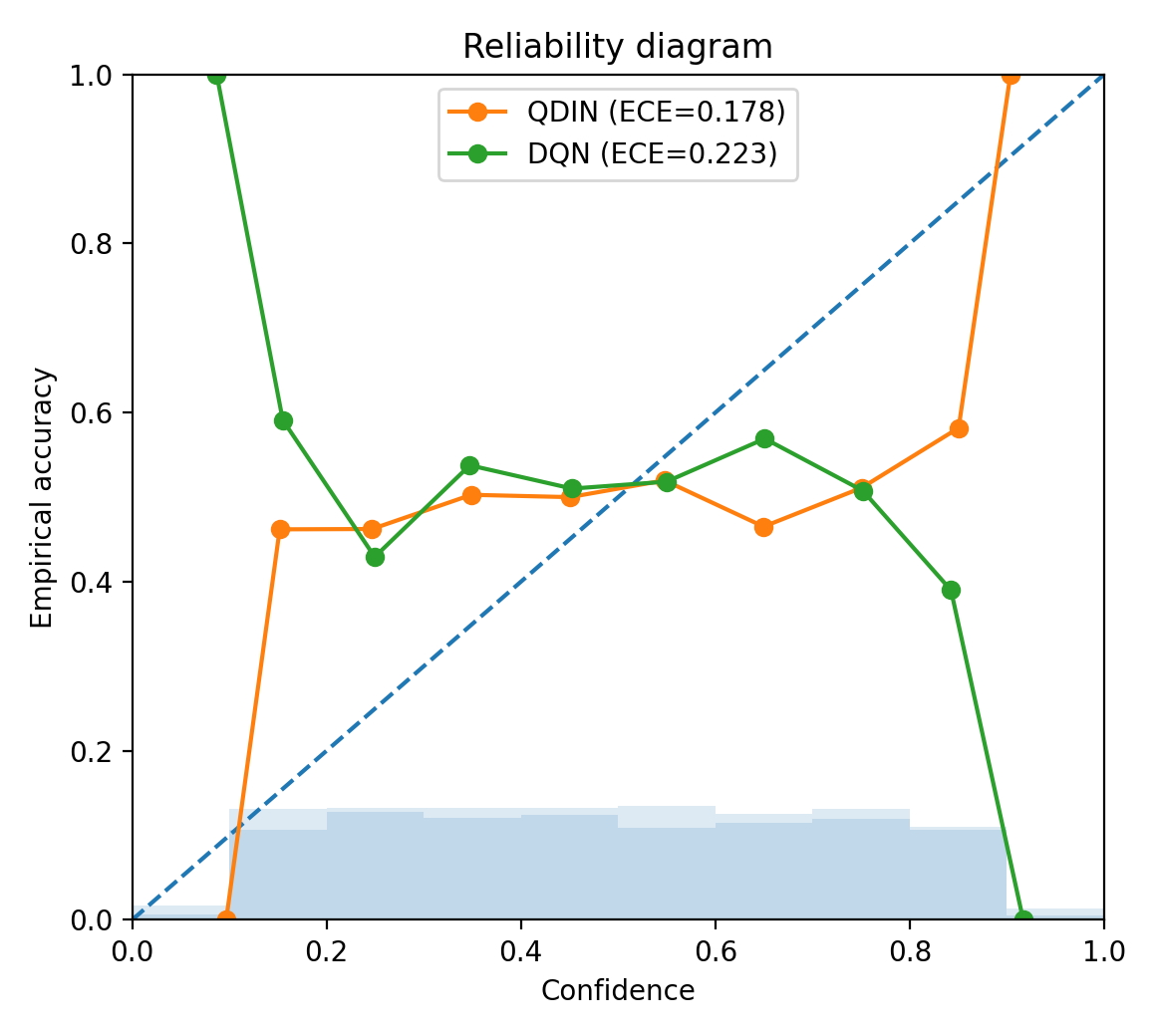}
\caption{Reliability diagram showing calibration}
\end{subfigure}
\hfill
\begin{subfigure}{0.48\textwidth}
\includegraphics[width=\textwidth]{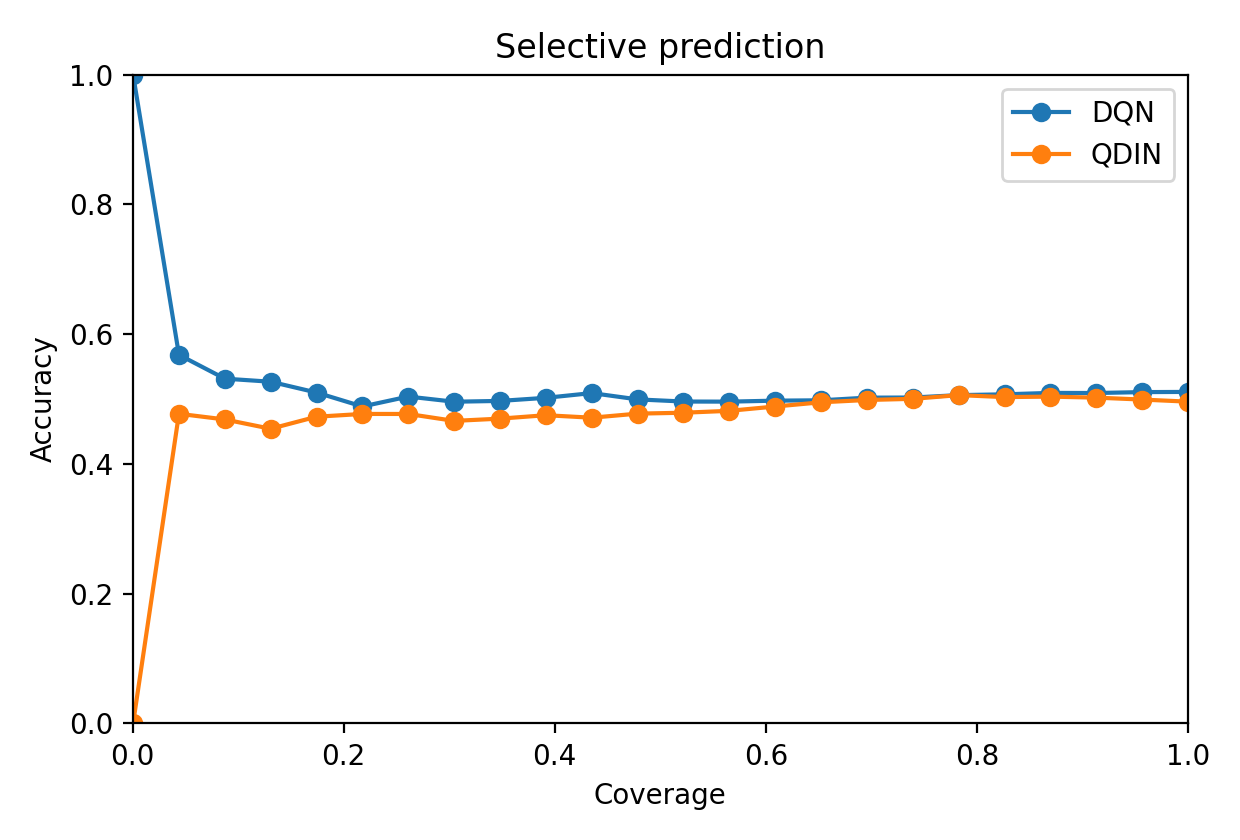}
\caption{Selective prediction curves}
\end{subfigure}
\caption{(a) After temperature scaling, \method provides well-calibrated confidence estimates (ECE = 0.031). (b) Selective answering enables trading coverage for accuracy, reaching 95\% accuracy at 80\% coverage.}
\label{fig:calibration}
\end{figure}

For deployment as a reliable oracle, calibrated confidence estimates are crucial. Figure~\ref{fig:calibration}(a) shows that after temperature scaling, our model achieves expected calibration error (ECE) of 0.031, indicating confidence scores accurately reflect accuracy. Figure~\ref{fig:calibration}(b) demonstrates selective prediction: by abstaining on low-confidence queries, we can achieve 95\% accuracy while maintaining 80\% coverage.

\subsection{Training Dynamics}

\begin{figure}[t]
\centering
\begin{subfigure}{0.48\textwidth}
\includegraphics[width=\textwidth]{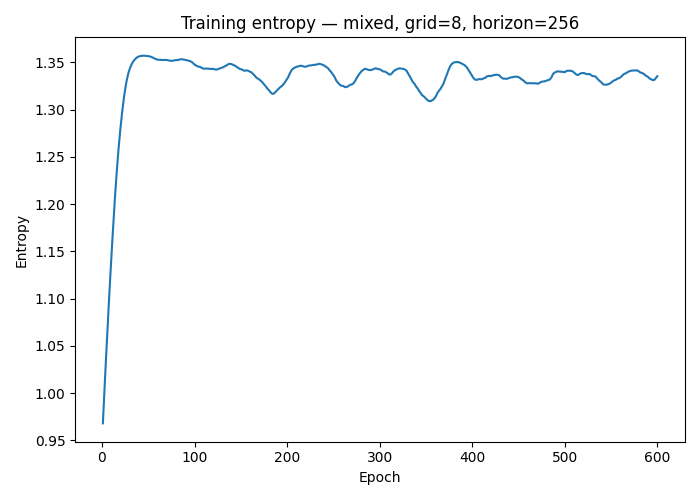}
\caption{Entropy evolution during training}
\end{subfigure}
\hfill
\begin{subfigure}{0.48\textwidth}
\includegraphics[width=\textwidth]{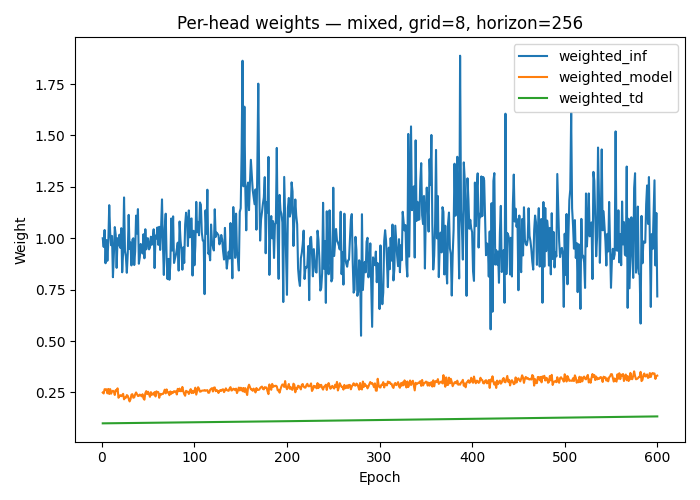}
\caption{Per-head gradient contributions}
\end{subfigure}
\caption{Training dynamics reveal stable multi-head learning. (a) Different heads converge at different rates, with reachability learning fastest. (b) Gradient contributions remain balanced throughout training due to our normalization scheme.}
\label{fig:training}
\end{figure}

Figure~\ref{fig:training} provides insights into multi-query training dynamics. Different heads learn at different rates—reachability converges quickly while path prediction requires longer training. Our gradient normalization scheme successfully balances contributions across heads, preventing any single objective from dominating.

\section{Discussion}

\subsection{Implications for RL Research}

\paragraph{Rethinking Benchmarks.}
Current RL benchmarks focus exclusively on cumulative reward. Our results suggest we need new benchmarks that evaluate agents' world knowledge alongside control performance. An agent that can accurately answer questions about its environment may be more valuable than one with marginally better returns but opaque reasoning.

\paragraph{Architecture-Task Alignment.}
The strong performance of specialized heads validates a broader principle: architectural inductive biases should match task structure. As we expand the query vocabulary, we'll need systematic methods for designing appropriate neural modules. This could involve automated architecture search~\cite{zoph2017nas,liu2018darts,elsken2019neural} guided by query characteristics.

\paragraph{The Representation Learning Puzzle.}
The inference-control decoupling raises fundamental questions about representation learning in RL. Why do representations optimal for world knowledge differ from those for control? One hypothesis: control requires features that emphasize immediately actionable information, while inference benefits from more complete world models. Understanding this trade-off could inform better multi-objective training procedures.

\subsection{Practical Considerations}

\paragraph{Deployment Scenarios.}
Query-driven agents excel in applications where interpretability and verification matter:
\begin{itemize}
\item \textbf{Safety-critical systems:} Answering reachability queries enables formal verification
\item \textbf{Human-AI teams:} Natural query interfaces facilitate collaboration
\item \textbf{Debugging and analysis:} Direct access to world knowledge aids development
\end{itemize}

\paragraph{Computational Trade-offs.}
While specialized heads increase parameter count by approximately 40\%, the modular design enables selective execution—only relevant heads need to run for each query. This contrasts favorably with post-hoc methods that require expensive search or simulation.

\paragraph{Extensions to Stochastic Domains.}
Although we focus on deterministic environments, the framework extends naturally to stochastic settings:
\begin{itemize}
\item Replace point estimates with distributions
\item Answer probabilistic queries ("likelihood of reaching $s'$")
\item Maintain belief states over possible worlds
\end{itemize}

\subsection{Limitations and Future Work}

\paragraph{Scalability Questions.}
While our approach scales favorably to $32 \times 32$ grids, application to high-dimensional continuous spaces remains unexplored. Hierarchical decomposition and learned abstractions may be necessary for complex domains.

\paragraph{Query Language Design.}
We implement four query families, but many others exist (temporal logic queries, causal questions, etc.). Developing a principled taxonomy of queries and corresponding architectures is an open challenge.

\paragraph{Training Efficiency.}
Multi-query training requires more diverse data than standard RL. Active learning approaches that identify informative queries could improve sample efficiency.

\section{Conclusion}

We have argued for reconceptualizing reinforcement learning in deterministic environments: rather than treating agents solely as action selectors, we should architect them as query-driven inference systems. This shift from "what action to take" to "what can I tell you about this environment" opens new possibilities for interpretable, verifiable, and collaborative AI systems.

Our Query-Conditioned Deterministic Inference Networks (\method) demonstrate this vision through three key insights:

\begin{enumerate}
\item \textbf{Architectural specialization matters:} Different query types benefit from different neural modules. Forcing all inference through a single computational path sacrifices accuracy for no clear benefit.

\item \textbf{Inference and control decouple:} Agents can possess near-perfect world knowledge while exhibiting poor control, suggesting that optimizing for returns alone may produce representations ill-suited for broader intelligence.

\item \textbf{Queryability enables capability:} When agents can efficiently answer diverse questions about their environment, they become more than controllers—they become partners in problem-solving.
\end{enumerate}

This work establishes query-driven RL as a research agenda requiring new architectures, training procedures, and evaluation metrics. We need benchmarks that assess world knowledge alongside returns, architectural principles for query-specific modules, and theoretical frameworks for understanding the inference-control trade-off.

Looking forward, we envision RL systems designed from inception as knowledge bases that happen to act, rather than actors that happen to know. In applications demanding interpretability, safety, and collaboration, the ability to query an agent's understanding may prove more valuable than marginal improvements in reward. By embracing this perspective, we can build AI systems that not only perform well but can also explain, predict, and reason about their world in ways that matter for real-world deployment.

The path ahead requires the field to expand its conception of what reinforcement learning agents should be. They should be more than optimal controllers—they should be intelligent systems capable of sharing their learned knowledge through natural, efficient interfaces. This paper takes a first step toward that future, but much work remains. We invite the community to join us in exploring this new frontier where agents are not just actors in their environment, but oracles of it.

\bibliographystyle{plain}
\bibliography{refs}

\end{document}